\documentclass{article}
\usepackage{ijcai22}

\usepackage[T1]{fontenc}
\usepackage{times}
\usepackage{soul}
\usepackage{url}
\usepackage[hidelinks]{hyperref}
\usepackage[utf8]{inputenc}
\usepackage[small]{caption}
\usepackage{graphicx}
\usepackage{amsmath}
\usepackage{amsthm}
\usepackage{amssymb}
\usepackage{amsfonts}
\usepackage{bm}
\usepackage{booktabs}
\usepackage{multirow}
\usepackage{algorithm}
\usepackage{algorithmic}
\usepackage{subcaption}
\usepackage{microtype}

\newcommand{\vect}[1]{\boldsymbol{\mathbf{#1}}}

\graphicspath{{figures/}}

\title{A Human-Centric Method for Generating Causal Explanations in Natural Language for Autonomous Vehicle Motion Planning}

\author{
Balint Gyevnar$^{1,}$\footnote{Contact author.}\and
Massimiliano Tamborski$^1$\and
Cheng Wang$^1$\and \\
Christopher G. Lucas$^1$\and
Shay B. Cohen$^1$\and 
Stefano V. Albrecht$^{1,2}$
\affiliations
$^1$School of Informatics, University of Edinburgh \\
$^2$Five AI Ltd., UK
\emails
\texttt{\{balint.gyevnar,cheng.wang,c.lucas,s.albrecht\}@ed.ac.uk, m.tamborski@sms.ed.ac.uk, scohen@inf.ed.ac.uk}
}

\begin{document}

\maketitle

\begin{abstract}
    Inscrutable AI systems are difficult to trust, especially if they operate in safety-critical settings like autonomous driving.
    Therefore, there is a need to build transparent and queryable systems to increase trust levels.
    We propose a transparent, human-centric explanation generation method for autonomous vehicle motion planning and prediction based on an existing white-box system called IGP2. 
    Our method integrates Bayesian networks with context-free generative rules and can give causal natural language explanations for the high-level driving behaviour of autonomous vehicles.
    Preliminary testing on simulated scenarios shows that our method captures the causes behind the actions of autonomous vehicles and generates intelligible explanations with varying complexity.
\end{abstract}

\section{Introduction}\label{sec:intro}
Autonomous vehicles (AVs) are predicted to improve, among other things, traffic efficiency and transport safety, reducing road fatalities possibly by as much as 90\%~\cite{wangEthicalDecisionMaking2022}.
AVs are also predicted to decrease pollution and make transportation more accessible for passengers with disabilities. 
However, the current complex, highly-integrated, and often opaque systems of AVs are not easily (or at all) understood by most humans.
This opaqueness often manifests in reluctance to accept the technology due to concerns that the vehicle might fail in unexpected situations~\cite{hussainAutonomousCarsResearch2019}.
This have fostered continued distrust and scepticism of AVs in the public eye~\cite{kimUsersTrustConcerns2021}.

We need to build trust in passengers if we want to overcome these psychological barriers and achieve wider acceptance for AVs. 
Crucial to the development of such trust, but neglected since the rise of black-box algorithms is the principle of explicability. This principle broadly means that the purposes, capabilities, and methods of the AV system must be \textbf{transparent}, that is, understandable and queryable by its passengers.
While this principle is generally important for any AI system, it is especially important for AVs as they operate in safety-critical settings and their decisions have far-reaching consequences on human lives.
There is a scientific consensus that we can increase transparency and build trust in AVs through the adoption of human-centric explainable AI (XAI)~\cite{hussainAutonomousCarsResearch2019,omeizaExplanationsAutonomousDriving2021,atakishiyevExplainableArtificialIntelligence2021}.
Humans prefer \textbf{causal} explanations~\cite{millerExplanationArtificialIntelligence2019}, so our explanations must be causally justifiable in terms of the processes that generated our actions.
We also want explanations to be \textbf{intelligible} for non-expert people to minimise cognitive overhead and help build general knowledge about AVs, reducing scepticism.
Finally, explanations must be faithful to the decision generating process to ensure they are not misleading.
We call this property the \textbf{soundness} of an explanation generation system.

\begin{figure}
    \centering
    \includegraphics[width=0.8\linewidth]{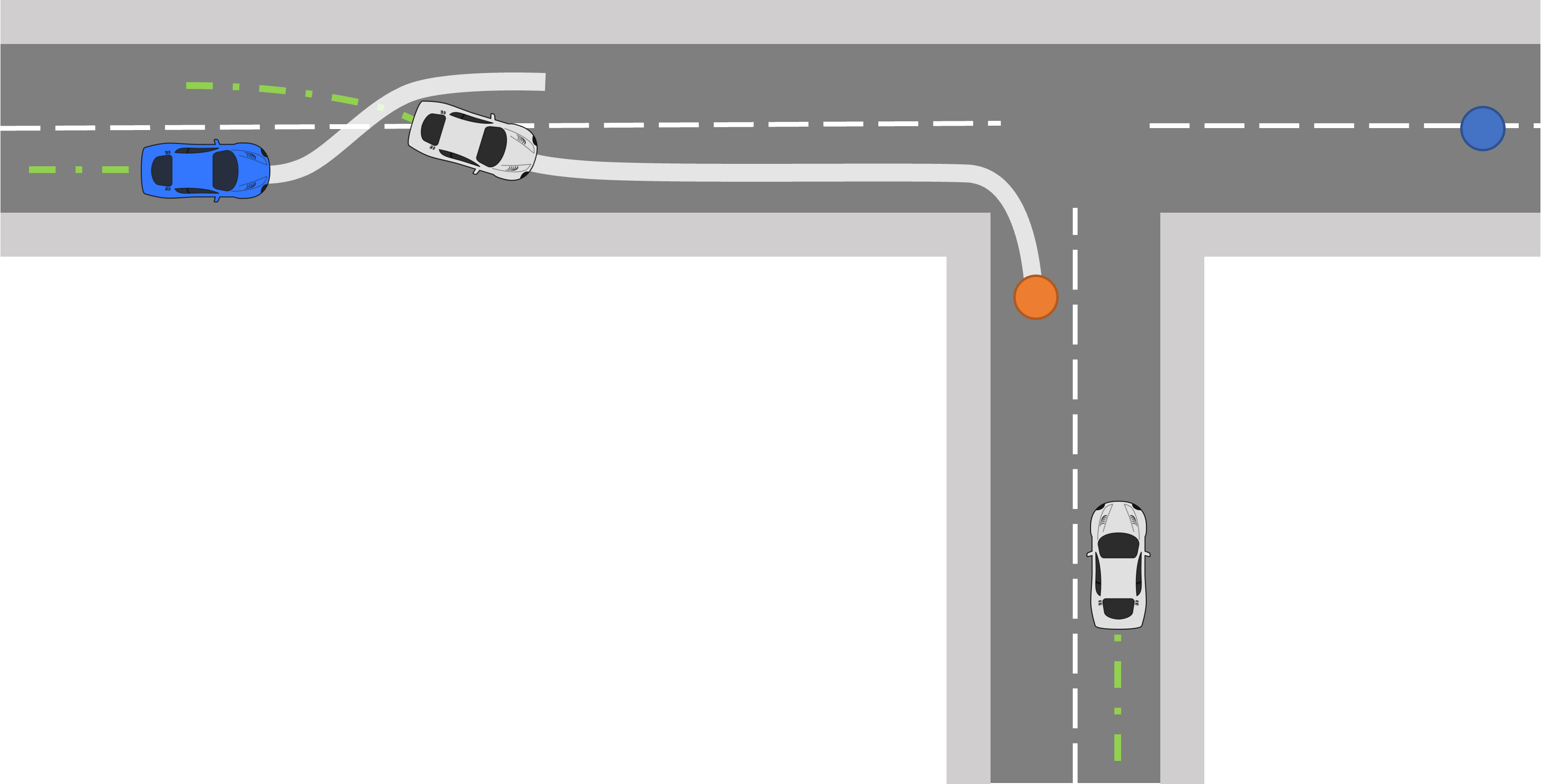}
    \caption{The ego vehicle (in blue) is heading straight to the blue goal but then changes lanes to the left. A passenger may inquire \textit{``Why did you change lanes instead of just driving straight?''}. Our system uses the motion planning and prediction calculations of the AV to give a causally justified contrastive explanation: if the ego had gone straight, then it would have likely reached its goal slower because the vehicle in front is probably changing lanes right and then exits right.}
    \label{fig:intro_example}
\end{figure}

We propose a human-centric \textit{explanation generation method} called eXplainable Autonomous Vehicle Intelligence (XAVI), focusing on high-level \textit{motion planning and prediction}.
XAVI is based on an existing transparent, inherently interpretable motion planning and prediction system called IGP2~\cite{albrechtInterpretableGoalbasedPrediction2021}.
An example of the output of our system is shown in Figure~\ref{fig:intro_example}.
Our method models the cause-effect relations behind the decisions of IGP2 as a Bayesian network allowing us to draw probabilistic causal judgements about the plans of the AV.
These judgements inform a context-free grammar that is used to generate intelligible \textit{contrastive explanations} (see Section~\ref{sec:background}), which compare the factual observed behaviour with some counterfactual behaviour.
Preliminary testing of XAVI on driving scenarios with baseline explanations by the authors of IGP2 demonstrates that our method correctly captures some of the causality behind the actions of the AV and generates intelligible natural language explanations with varying complexity.
We end with a discussion outlining important future work for explaining AV behaviour.
We also release the code for XAVI on GitHub\footnote{\url{https://github.com/uoe-agents/xavi-ai4ad}}.

\section{Background}\label{sec:background}

Most existing methods of XAI are model-agnostic approaches that focus on classification/regression tasks relying on black-box and surrogate models. 
Local surrogate models usually calculate some feature importance ordering given a \textit{single input} instance for a given black box model~\cite{ribeiroWhyShouldTrust2016,lundbergUnifiedApproachInterpreting2017,montavonExplainingNonlinearClassification2017}.
However, instance-based explanations may ignore the overall workings of the black box and, when interpreted incorrectly, may build an incorrect understanding of our system in humans.
Global surrogate models instead use black box models as a teacher to train simpler, interpretable white box systems~\cite{bastaniInterpretingBlackboxModels2017,lakkarajuInterpretableExplorableApproximations2017}.
These can model the overall decision process of the teacher, though usually at the cost of soundness.
In general, surrogate approaches have to introduce a new layer of abstraction that does not allow or distort the causal understanding of the decision process of the underlying black box and may introduce unwanted biases to the explanations.
In addition, the output of these methods is usually expert oriented and difficult to understand for non-experts.

These shortcomings motivated several recent work that popularise a human-centric approach to explanation generation based on causality and intelligibility~\cite{millerExplanationArtificialIntelligence2019,dazeleyLevelsExplainableArtificial2021,ehsanHumancenteredExplainableAI2020}.
In the case of classical AI planning, XAI-PLAN~\cite{borgoProvidingExplanationsAI2018} answers contrastive questions of the form ``Why do X instead of Y?'', while WHY-PLAN \cite{korpanNaturalExplanationsRobot2018} generates natural language explanations based on model reconciliation, which compares the generated plan of the system to a user-given alternative plan.
These methods represent a shift towards a more human-centric approach, however the main issue with classical AI planning methods is their reliance on fixed domain descriptions which make their use in dynamic and complex environments such as autonomous driving difficult.

Furthermore, while the motivation for building trust and transparency for AVs is well understood, few works have proposed methods that use AV domain knowledge to inform their explanation generation system.
Previous methods used deep learning to generate textual explanations for AVs based on a data set of annotated recordings with textual explanations called BDD-X~\cite{kimTextualExplanationsSelfDriving2018,ben-younesDrivingBehaviorExplanation2022}.
Additionally, \citeauthor{omeizaAccountabilityProvidingIntelligible2021}~(\citeyear{omeizaAccountabilityProvidingIntelligible2021}) proposed an explanation generation system based on decision trees taught by a black box and using language templates.
These methods generate intelligible explanations, but the generating processes are surrogate models which are neither causal nor transparent.

Recently, \citeauthor{albrechtInterpretableGoalbasedPrediction2021}~(\citeyear{albrechtInterpretableGoalbasedPrediction2021}) proposed an inherently interpretable integrated planning and prediction system called IGP2.
This method relies on intuitive high-level driving actions and uses rational inverse planning to predict the trajectories of other vehicles, which are then used to inform motion planning with Monte Carlo Tree Search (MCTS).
In this work, we rely on IGP2 as it is a white-box model, whose internal representations can be directly accessed while its decisions can be readily interpreted through rationality principles.
Direct access to internal representations means access to the MCTS tree search which naturally allows for causal interpretation.
We directly leverage this inherent causality to build our method.

\section{IGP2: Interpretable Goal-Based Prediction and Planning for Autonomous Driving}\label{sec:igp2}

\begin{figure*}
    \centering
    \includegraphics[width=\textwidth]{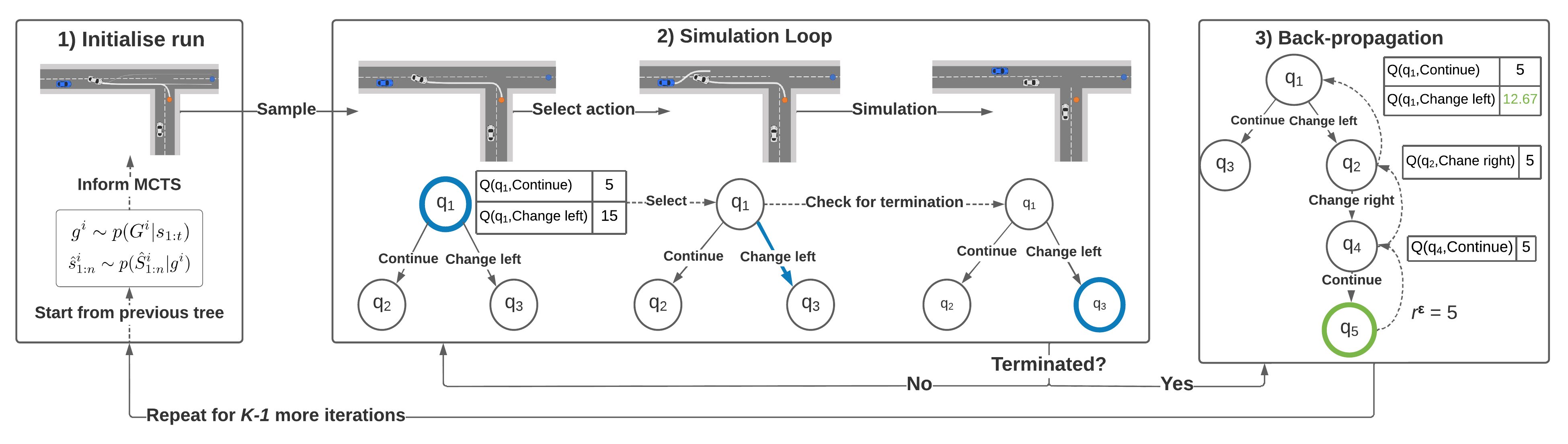}
    \caption{
    MCTS at work: \textit{(Step 1)} Before each simulation, we sample and fix the trajectories of each non-ego vehicle.
    \textit{(Step 2)} From the current state (in blue) we select our next macro action based on Q-values. In this example, we selected \textit{Change-left}, which is then forward simulated while the other traffic participants follow their fixed trajectories. During simulation we check for termination conditions. \textit{(Step 3)} If the ego reached a goal, or some other termination condition was met (e.g. collision), the ego receives a reward $r^\varepsilon$ which is back-propagated through the trace of macro actions that reached the termination state to update the Q-values of each action. \textit{(Step 4)} We repeat the process until $K$ iterations are reached resulting in a search tree with maximal depth $d_{max}$.}
    \label{fig:mcts_igp2}
\end{figure*}

In the following, we give a brief introduction to the notation and methods of IGP2.
Let $\mathcal{I}$ be the set of traffic participants in the local neighbourhood of the ego vehicle  denoted $\varepsilon \in \mathcal{I}$.
At time step $t$ each traffic participant $i \in \mathcal{I}$ is in a local state $s^i_t \in \mathcal{S}^i$ which includes its pose (position and heading), velocity, and acceleration. 
The joint state of all vehicles is denoted $s_t \in \mathcal{S} = \times_i \mathcal{S}^i$, and the state sequence $(s_a, \dots, s_b)$ is written $s_{a:b}$.
A trajectory is defined as a state sequence, where two adjacent states have a time step difference of one.
IGP2 is goal-oriented, so it assumes that each traffic participant is trying to reach one of a finite number of possible goals $g^i \in \mathcal{G}^i$. 
Trajectories can be used to calculate rewards $r^i$ for a vehicle, where $r^i$ is the weighted linear sum of reward components corresponding to aspects of the trajectory.
The set of reward components is denoted by $\mathcal{C}$ and consists of \textit{time-to-reach-goal}, \textit{jerk}, \textit{angular acceleration}, \textit{curvature}, \textit{collision}, and \textit{termination} (received when IGP2 runs out of computational budget).
Some reward components are mutually exclusive.
For example, if we receive a (negative) ``reward'' for collision, then we cannot receive a reward for anything else.

The planning problem of IGP2 is to find an optimal policy for the ego vehicle $\varepsilon$ that selects actions given a joint state to reach its goal $g^\varepsilon$ while optimising its reward $r^\varepsilon$. 
Instead of planning over low-level controls, IGP2 defines higher-level manoeuvres with applicability and termination conditions, and (if applicable) a local trajectory $\hat{s}^i_{1:n}$ for the vehicle to follow.
IGP2 uses the following manoeuvres: \textit{lane-follow}, \textit{lane-change-\{left,right\}}, \textit{turn-\{left,right\}}, \textit{give-way}, and \textit{stop}. 
These manoeuvres are then further chained together into macro actions denoted here with $\omega\in\hat{\Omega}$, which are common sequences of manoeuvres parameterised by the macro actions. 
The set of all macro-actions is $\hat{\Omega}=\{$\textit{Continue}, \textit{Change-\{left,right\}}, \textit{Exit}, \textit{Continue-next-exit}, \textit{Stop}$\}$.
IGP2 searches for the optimal plan over these macro actions.

Finding the optimal plan for the ego vehicle has two major phases. 
First in the \textit{goal and trajectory recognition phase} (referred to as goal recognition from here on), IGP2 calculates a distribution over goals $G^i\subseteq\mathcal{G}^i$ given the already observed trajectory of vehicle $i$ denoted $p(G^i|s^i_{1:t})$.
To each goal we then generate a distribution over possible trajectories ${S}^i_{1:n}\subseteq{\mathcal{S}}^i_{1:n}$ given by $p({S}^i_{1:n}|G^i)$.
In the \textit{planning phase}, goal recognition is used to inform a Monte Carlo Tree Search (MCTS) algorithm over macro actions, which finds the optimal sequence of macro actions (i.e. plan) by simulating many possible plans over $K$ iterations to see how each plan interacts with the other traffic participants.
More details of MCTS can be found in the caption of Figure~\ref{fig:mcts_igp2}.
During planning we track and accumulate all relevant information about how decisions are made which we then use to initialise our explanation generation system.

\section{eXplainable Artificial Vehicle Intelligence}\label{sec:xavi}

\begin{figure}[t]
    \centering
    \includegraphics[width=\linewidth]{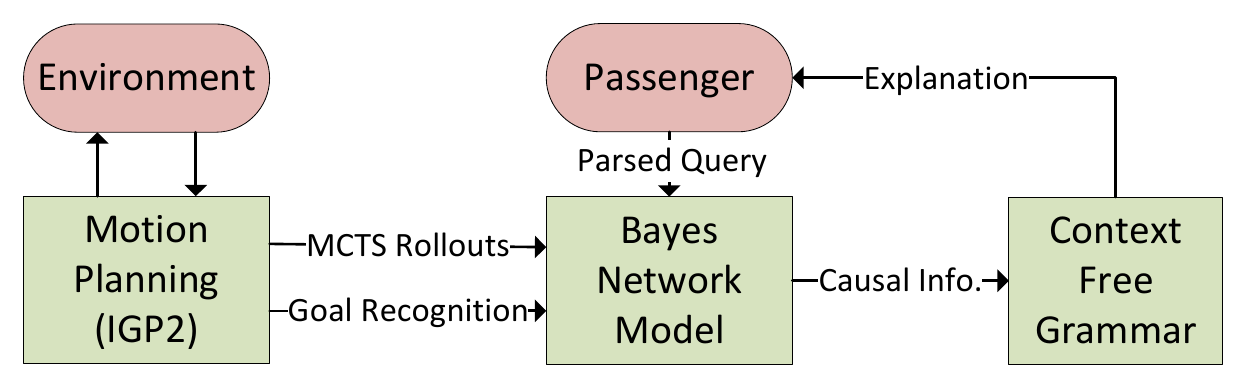}
    \caption{The XAVI system. IGP2 interacts with its environment and generates an optimal plan using MCTS and predictions from goal recognition. The accumulated information about how these components arrived at the optimal plan is used to build a Bayesian network (BN) model. We compare this model to a contrastive query from the passenger and extract causal relationships, which are fed to a context-free grammar that generates natural language explanations.}
    \label{fig:xavi}
\end{figure}

Though IGP2 is transparent and can be readily interpreted, it requires expert domain-knowledge to interpret its results and present intuitive explanations that are intelligible for the non-expert passenger.
It is therefore desirable to automate the interpretation and explanation procedure in the human-centric way we outlined in Section~\ref{sec:intro}.

We therefore present our \textit{explanation generation system} called eXplainable Artificial Vehicle Intelligence (XAVI).
The overall architecture of XAVI can be seen in Figure~\ref{fig:xavi}.
The core idea of XAVI is to map the accumulated information about goal recognition and the steps of a complete MCTS planning run to random variables, which we then use to construct a Bayesian network (BN) model that encodes the properties of that particular MCTS planning run. 
This allows us to derive probabilistic causal information about alternative (i.e. counterfactual) sequences of macro actions and their possible effects on rewards and outcome.

Counterfactuals are a crucial part of XAVI, as the generated explanations \textit{contrast} the factual, optimal plan with a counterfactual plan in which the ego would have followed a different sequence of macro actions.
Contrastive explanations are studied in philosophical literature where most argue that all \textit{why}-questions are (implicitly) contrastive~\cite{millerExplanationArtificialIntelligence2019}.
This means that our generated explanations have a form similar to: ``\textit{If we had done <CF> [instead of <F>], then <EFFECTS> would have happened, because <CAUSES>.}''. 
Here \textit{<F>} and \textit{<CF>} are the factual and counterfactual macro actions respectively, while \textit{<EFFECTS>} are the changes to reward components and outcome in the counterfactual scenario.
\textit{<CAUSES>} describe relevant features of the trajectories of traffic participants (including the ego) that have caused the changes in \textit{<EFFECTS>}.
Note, we omit explicitly mentioning the factual \textit{<F>} in our explanations since we assume that the passenger observed the ego's actions and is aware of what actions the ego had taken.
We also assume, that the passenger's query is in a parsed format that allows us to directly extract counterfactual causal information from our Bayesian network model.

    \subsection{Bayes Network Model}\label{sec:xavi:bn}

        \subsubsection{Random Variables}
        The first step to create the Bayesian network model is to map MCTS planning steps to random variables.
        MCTS starts by sampling goals and trajectories for each non-ego vehicle $i$.
        Let the vector of random variables corresponding to goal sampling (we are not sampling for the ego) be $\vect{G}=[G^1,\dots,G^{|\mathcal{I}|-1}]$ and the vector of trajectories be $\vect{S}=[S^1,\dots,S^{|\mathcal{I}|-1}]$.
        The values of $G^i\in\mathcal{G}^i$ and $S^i\in\mathcal{S}^i_{1:n}$ are from the set of possible goals and trajectories for vehicle $i$.
        For example, setting $G^i=g^i$ means that we sample goal $g^i$ for $i$.

        Next, for every macro action selection step in the MCTS search tree, that is for each depth $1\leq d\leq d_{max}$ in the tree, we define a random variable $\Omega_d$ with support of $\hat{\Omega}_d\subseteq\hat{\Omega}$ which is the set of all applicable macro actions at depth $d$.
        Each $\Omega_d$ may also take the value of the empty set $\varnothing$, which means that no action was selected at depth $d$.
        We collect each of these random variables into a single vector denoted $\vect{\Omega}=[\Omega_1,\dots,\Omega_{d_{max}}]$.
        This means that specifying a trace in the search tree corresponds to assigning an action to each $\Omega_d$ which we can represent as a vector $\vect{\omega}=[\omega_1,\dots,\omega_{d_{max}}]$.

        For each reward component $c\in\mathcal{C}$ that MCTS uses to calculate $r^\varepsilon$ we can define a continuous random variable $R_c\in\mathbb{R}$ that gives the value for that particular reward component, or is $\varnothing$ if the reward component is not present.
        For each $R_c$, let $R^b_c\in\{0,1\}$ be a binary variable that indicates the existence of reward component $c$.
        That is, if $R^b_c=1$ then $R_c\neq\varnothing$.
        Let the vectors that collect the random variables for each component be $\vect{R}=[R_{c}]_{c\in\mathcal{C}}$ and similarly for $\vect{R}^b$.

        Finally, we define outcome variables.
        It is important to note, that IGP2 does not explicitly represent various types of outcomes, so these variables do not correspond to any actual steps in MCTS.
        Instead, outcome variables are used here to conveniently describe the state of the ego vehicle at the termination of a simulation.
        There are four outcome types given by the set $\mathcal{O}$: \textit{done} ($g^\varepsilon$ was reached), \textit{collision}, \textit{termination} (reached $d_{max}$ in MCTS without reaching $g^\varepsilon$), and \textit{dead} (for any outcomes not covered by the previous three).
        For each outcome type $k\in\mathcal{O}$ we define a corresponding binary variable $O_k\in\{0,1\}$ which indicates whether that outcome was reached at the termination of a MCTS simulation.
        The vector of outcome variables is denoted with $\vect{O}$.

        \subsubsection{Joint Factorisation}

        \begin{figure}[t]
            \centering
            \includegraphics[width=\linewidth]{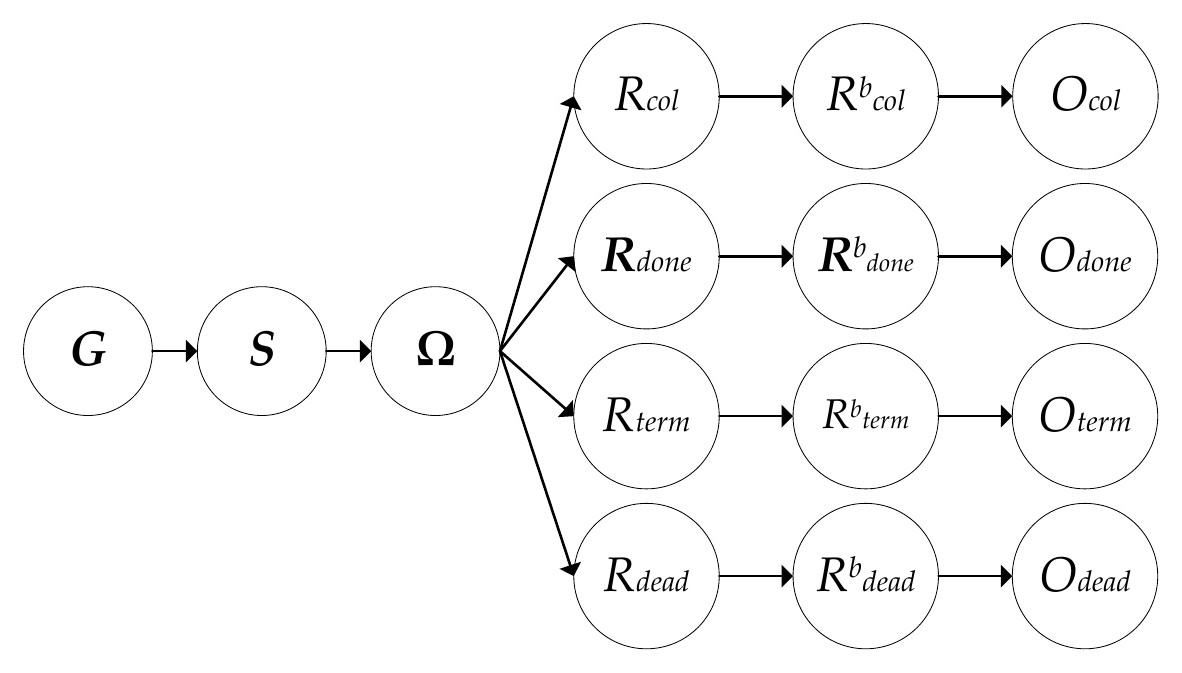}
            \caption{The underlying DAG of the Bayes network used for factoring the joint over all random variables. The chain-rule for $\vect{\Omega}$ is not shown explicitly. The associated reward components for $O_{done}$ are $\vect{R}_{done}\in\{time, jerk, angular\text{-}acceleration, curvature\}$.}
            \label{fig:mcts_bn}
        \end{figure}

        We now define the directed acyclic graph (DAG) to factorise the joint over random variables defined in the previous section and describe the probability distributions of each factor.

        Consider the DAG shown in Figure~\ref{fig:mcts_bn}. 
        Goals for non-ego vehicles are sampled independently according to their distributions from goal recognition while the trajectories depend on the sampled goals.
        The joint distribution of these variables over each vehicle except the ego (hence iterating over $i \in \mathcal{I} \setminus \varepsilon$) are given below, which simply state that the probabilities of goals and trajectories of vehicles are mutually independent:
        \begin{align}
            p(\vect{G}|\vect{s}_{1:t})&=\prod_{i\in\mathcal{I}\setminus\varepsilon}p(G^i|s^i_{1:t}), \label{eq:pg} \\
            p(\vect{S}|\vect{G})&=\prod_{i\in\mathcal{I}\setminus\varepsilon}p(\hat{s}^i_{1:n}|G^i). \label{eq:ps_g}
        \end{align}

        Because trajectories of other traffic participants affect what macro actions are selected in MCTS, the random variables $\vect{\Omega}$ are conditioned on $\vect{S}$.
        Furthermore, the joint distribution of macro action selections $\vect{\Omega}$ is given by the chain rule, which corresponds to the product of macro action selection probabilities in the MCTS tree along a search trace:
        \begin{equation}\label{eq:omega_s}
            p(\vect{\Omega}|\vect{S})=p(\Omega_1|\vect{S})\prod_{d=2}^{d_{max}}p(\Omega_{d}|\vect{\Omega}_{1:d-1},\vect{S}).
        \end{equation}
        The definition of $p(\Omega_{d}|\vect{\Omega}_{1:d-1},\vect{S})$ corresponds to the probabilities of selecting a macro action at depth $d$ from the unique state $s$ reached by following $\vect{\Omega}_{1:d-1}$.
        For each value $\omega_d\in\Omega_d$ we estimate $p(\omega_d|\vect{\Omega}_{1:d-1},\vect{S})$ from the $K$ simulations of MCTS as the number of times $\omega_d$ was selected in state $s$ over the total number of times any action was selected in state $s$ (i.e. the total number of visits of state $s$).

        Reward components in $\vect{R}$ depend on the driven trajectory of the ego, and therefore the selected sequence of macro actions given by $\vect{\Omega}$.
        However components are otherwise calculated independently from one another.
        The joint distribution for $\vect{R}$ is the product of distributions of each reward component:
        \begin{equation}\label{eq:pr_omega}
            p(\vect{R}|\vect{\Omega})=\prod_{c\in\mathcal{C}}p(R_c|\vect{\Omega}).
        \end{equation}
        If $s$ now denotes the state reached by following $\vect{\Omega}$, we estimate $p(R_c|\vect{\Omega})$ from the $K$ simulations as a normal distribution with sample mean $\mu_c(s)$ and sample variance $\sigma^2(s)$ calculated from the values observed for $R_c$ in state $s$.

        The existence indicator variables depend only on their corresponding reward component and their joint distribution otherwise assumes mutual independence:
        \begin{equation}\label{eq:prb_r}
            p(\vect{R}^b|\vect{R})=\prod_{c\in\mathcal{C}}p(R^b_c|R_c),
        \end{equation}
        \noindent where $p(R^b_c=1|R_c)=1$ iff $R_c\neq\varnothing$, so $R^b_c=1$ only when we have observed some non-empty value for $R_c$.

        Finally, the outcome variables depend only on the existence of certain reward components as given in Figure~\ref{fig:mcts_bn}.
        For each outcome variable $k\in\mathcal{O}$ let $\vect{R}^b_k\subseteq \vect{R}^b$ be the vector of random variables of binary reward components that $k$ depends on.
        The joint distribution of outcomes is mutually independent:
        \begin{equation}\label{eq:po_rb}
            p(\vect{O}|\vect{R}^b)=\prod_{k\in\mathcal{O}}p(O_k|\vect{R}^b_k),
        \end{equation}
        \noindent where $p(O_k=1|\vect{R}^b_k)=1$ iff every reward component $R^b_c\in\vect{R}^b_t$ is not $\varnothing$.
        That is, the outcome $k$ is true iff all corresponding reward components have been observed at the termination of the MCTS simulation.

        Finally, by multiplying the left-hand side of Equations~\ref{eq:pg}-\ref{eq:po_rb} we get the joint distribution over all random variables.
        The binary random variables $\vect{R}^b$ are primarily used to simplify the calculation of the outcome probability distribution over $\vect{O}$, so for most calculation we marginalise $\vect{R}^b$ out, giving the joint we work with: $p(\vect{G,S,\Omega,R,O})$.

        \subsubsection{Note on Complexity}\label{sec:bn:complexity}
        The size of the conditional probability distributions (CPDs) for $p(\omega_d|\vect{\Omega}_{1:d-1},\vect{S})$ can, in the worst case, grow exponentially with the depth $d$ according to $\mathcal{O}(|\hat{\Omega}|^d)$.
        However, there are two reasons why this is not a prohibitive issue.
        First, the search trees of IGP2 are very shallow, usually $d_{max}\leq 4$, and secondly since the MCTS tree is sparse, most values of the CPDs are zero.
        So instead of storing the full CPDs explicitly, we can associate each CPD to the state it is applicable in (given by $\vect{\Omega}_{1:d-1}$) and calculate the needed probabilities on-the-fly.

    \subsection{Extracting Causal Information}
    From the joint distribution we can infer various conditional distributions that allow us to draw causal judgements about counterfactual scenarios.
    Let us assume that MCTS selected the factual, optimal plan for the ego denoted with $\vect{\omega}_F=[\omega_1,\dots,\omega_{d_{max}}]$.
    Further assume, that the passenger query describes a (possibly incomplete) set of counterfactual macro actions $\vect{\omega}_{CF}=[\omega_{j_1},\dots,\omega_{j_n}]$ corresponding to the random variables $\Omega_{j_1},\dots,\Omega_{j_n}$ indexed by the set $\mathcal{J}=\{j_1,\dots,j_n\}$.

    First, we calculate the outcome distribution of $\vect{O}$ given the counterfactual, that is the distribution:
    \begin{equation}\label{eq:po_cf}
        p(\vect{O}|\vect{\omega}_{CF}).
    \end{equation}
    This allows us to determine how the outcome of MCTS would have changed if ego had followed the counterfactual actions.

    Second, we want to determine how the reward components differ from the factual to the counterfactual scenario.
    This would allow us to order the components by the amount that they were affected by the switch to $\vect{\omega}_{CF}$, and we can use this ordering to populate the \textit{<EFFECTS>} variable in our explanation.
    Formally we can do this, by calculating:
    \begin{equation}\label{eq:delta_r}
        \Delta\vect{R}=\mathbb{E}[\vect{R}|\vect{\omega}_{F}] - \mathbb{E}[\vect{R}|\vect{\omega}_{CF}].
    \end{equation}
    We can sort $\Delta\vect{R}$ in decreasing order by the absolute value of its elements to get the required ordering.

    Finally, we would like to determine how much the trajectories of each individual non-ego participant affect the macro action selections of the ego.
    We can use this information to determine which traffic participants are most relevant to mention in our explanations and in what order. 
    We can derive this ordering by comparing how the marginal distribution of macro actions $p(\vect{\Omega})$ changes when conditioned on different trajectories of non-egos. 
    Since $p(\vect{\Omega})$ already encodes the optimal sequence of macro actions taking into account the trajectories of other participants, we are trying to find the conditional distribution that changes the marginal the least.
    Formally, for a vehicle $i$ and for each of its possible goals $g\in G^i$ and trajectories $s\in S^i_{1:n}$ we calculate the Kullback-Leibler divergence between the marginal and the conditional of $\vect{\Omega}$:
    \begin{equation}\label{eq:dkl}
        \begin{split}
            D^i_{g,s}&=D_{KL}[p(\vect{\Omega})||p(\vect{\Omega}|G^i=g,S^i=s)] \\
            &=\sum_{\vect{\omega}\in\vect{\Omega}} p(\vect{\omega})\log_2\left(\frac{p(\vect{\omega})}{p(\vect{\omega}|g,s)}\right).
        \end{split}
    \end{equation}
    If $D^i_{g,s}$ is the same for all goals and trajectories, that implies that the actions of vehicle $i$ does not affect the actions of the ego, so we will ignore vehicle $i$.
    Otherwise, we can sort all $D^i_{g,s}$ increasingly giving as an ordering on the importance of vehicles, goals, and trajectories.
    Note, that if vehicle $i$ has only a single predicted goal and trajectory then $D^i_{g,s}=0$.
    In this case we cannot use this measure to determine whether the vehicle $i$ interacted with the ego or not.
    A more robust method to replace this measure would be to repeat the MCTS planning with each vehicle $i$ removed from the simulations and looking at whether the actions of the ego have changed.
    This may however be computationally quite expensive to do.

    \subsection{Generating Natural Language Explanations}
    To generate intelligible explanations from the information derived in the previous section, we define a set of (recursive) generative rules given by a context-free grammar.
    We feed the extracted information to this grammar, which will generate a unique sentence.
    Since the raw generated sentences may be somewhat unnatural, we apply a post-processing step, where commonly occurring complex expressions are converted to simpler phrases (e.g. with higher time to goal $\rightarrow$ slower).

    The complete set of generative rules is given in the appendix in Figure~\ref{fig:cfg}.
    To instantiate this grammar we pass the following information to it:
    \begin{itemize}
        \item $s$: Information about the counterfactual scenario containing three fields: the counterfactual macro actions $s.\vect{\omega}$, as well as the most likely outcome $s.o$ and its probability $s.p$ as given by Equation~\ref{eq:po_cf}.
        \item $\vect{e}$: A list of effects on reward components of switching to the counterfactual. Each element $e\in\vect{e}$ contains two fields: the difference in reward $e.\delta$ as given by Equation~\ref{eq:delta_r}, and the name of the reward component $e.r$ corresponding to the difference.
        \item $\vect{c}$: A list of causes that resulted in the effects we observed. Each cause $c\in\vect{c}$ has three fields: the non-ego traffic participant $c.i$ the cause is related to, the trajectory (and the macro actions that generated it) $c.\vect{\omega}$ the non-ego is taking as calculated using Equation~\ref{eq:dkl}, and the probability $c.p$ of the ego taking that trajectory.
    \end{itemize}

    For example, assume that we give to the CFG the following: $s=\{\vect{\omega}:[\text{\textit{Continue}}],o:\text{\textit{done}},p:0.75\}$.
    We also have $\vect{e}=[\{\delta:-5,r:\text{time}\}]$, and finally we got $\vect{c}=[\{i:1,\vect{\omega}:[\text{\textit{Change-right}}],p:0.6\}]$. Then the generated explanation before post-processing would be:
    \textit{``If ego had continued ahead then it would have likely reached its goal with lower time to goal because vehicle 1 would have probably changed right.''}.

\section{Experiments}\label{sec:experiment}

\begin{figure*}
    \centering
    \begin{subfigure}[b]{0.33\textwidth}
        \includegraphics[width=\linewidth]{s1-figure.png}
    \end{subfigure}
    \hspace{13px}
    \begin{subfigure}[b]{0.33\textwidth}
        \includegraphics[width=\linewidth]{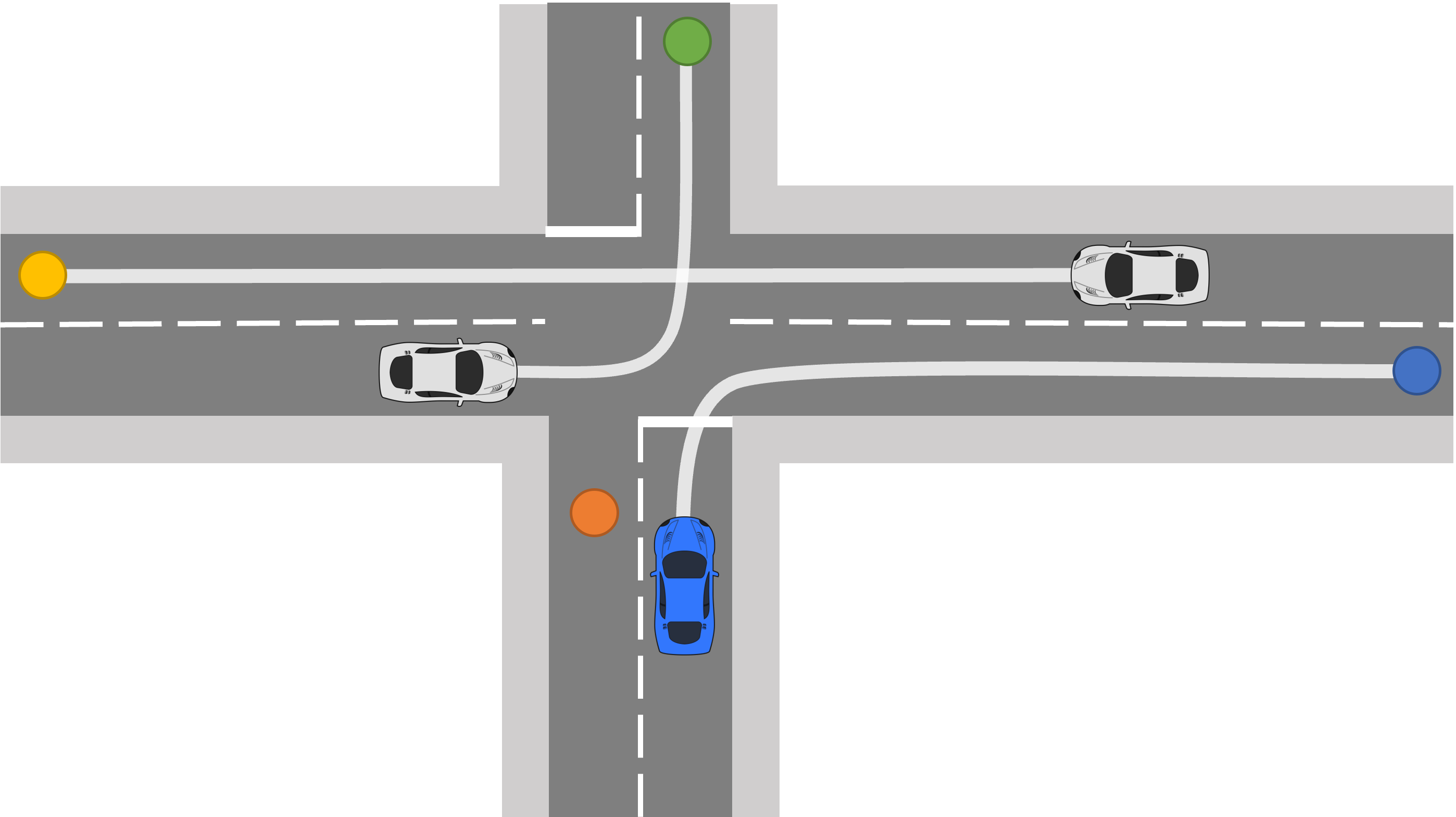}
    \end{subfigure}
    \hspace{5px}
    \caption{\textit{(Left; S1)}. The ego vehicle (in blue) starts out in the right lane with its goal being to reach the end of the road it is currently on. The vehicle in front of the ego (vehicle 1) starts in the left lane. At one point vehicle 1 cuts in front of the ego by changing lanes right and then begins slowing down. This behaviour is only rational if vehicle 1 intends to turn right at the junction ahead. To avoid being slowed down, the ego changes to the left lane. The factual, optimal actions of the ego in this scenario are therefore $\vect{\omega}^{1}_F=[\text{\textit{Change-left}},\text{\textit{Continue}}]$.  \textit{(Right; S2)} The ego is trying to turn right and approaches the junction. Ego sees the vehicle on its left on a priority road (vehicle 1) slow down for a stop. Considering that there is an oncoming vehicle from the right coming at high speed (vehicle 2) the action of vehicle 1 is only rational if its goal is to turn left and it is stopping to give way. Ego can use the time while vehicle 1 is stopped to turn right earlier, instead of waiting until vehicle 1 passes. The factual actions of ego is: $\vect{\omega}^{2}_F=[\text{\textit{Exit-right}},\text{\textit{Continue}}]$. }
    \label{fig:scenarios}
\end{figure*}

\begin{table*}[]
    \centering
    \begin{tabular}{lp{40em}}
         $\vect{\omega}^1_{CF}$ & \textbf{Generated Explanation} \\
         \hline
         \multirow{3}{*}{\textit{Continue}} & \textbf{\textit{``If ego had gone straight then it would have$\dots$''}} \\
         & \textit{``likely reached the goal slower because vehicle 1 probably changes right then exits right.''} \\
         & \textit{``likely collided with vehicle 1 because vehicle 1 probably changes right then exits right.''}\\
         \hline
         \multirow{3}{*}{\textit{\textit{Exit-right}}} & \textbf{\textit{``If ego had turned right then it would have$\dots$''}} \\
         & \textit{``not reached the goal.''} \\
         & \textit{``collided with vehicle 1 because vehicle 1 probably changes right and exits right.''}
    \end{tabular}
    \caption{\textit{(Scenario S1)} Explanations with one cause and one effect. Our system can successfully determine the cause and effect of the lane change and the effect of exiting right. 
    Note, that the system also captures the possibility of collisions when the ego and vehicle 1 start close to one another, as in this case ego cannot break quickly enough to avoid vehicle 1 cutting in front of it.}
    \label{tab:s1_expl}
\end{table*}

\begin{table*}[]
    \centering
    \begin{tabular}{lp{40em}}
         $\vect{\omega}^2_{CF}$ & \textbf{Generated Explanation} \\
         \hline
         \multirow{3}{*}{\textit{Exit-straight}} & \textbf{\textit{``If ego had gone straight then it would have$\dots$''}} \\
         & \textit{``not reached the goal.''} \\
         & \textit{``not reached the goal because vehicle 1 likely turns left.''}\\
         \hline
         \multirow{3}{*}{\textit{\textit{Exit-left}}} & \textbf{\textit{``If ego had turned left then it would have$\dots$''}} \\
         & \textit{``not reached the goal.''} \\
         & \textit{``not reached the goal because vehicle 1 likely turns left.''}
    \end{tabular}
    \caption{\textit{(Scenario S2)} Explanations with at most one cause and one effect. Both counterfactuals result in non-completion of the ego's goal, which is correctly captured as well as the rational action of vehicle 1 turning left. Note, the vehicle described in the causes affects the motion of the ego but is not directly responsible for the \textit{counterfactual} outcome of the ego not reaching its goal.}
    \label{tab:s2_expl}
\end{table*}

\begin{table*}[]
    \centering
    \begin{tabular}{lp{40em}}
         $\vect{\omega}^1_{CF}$ & \textbf{Generated Explanation} \\
         \hline
         \multirow{6}{*}{\vspace{-0em}\textit{Continue}} & \textbf{\textit{``If ego had gone straight then it would have$\dots$''}} \\
         & \textit{``likely reached the goal slower because vehicle 1 probably changes right then exits right.''} \\
         & \textit{``likely reached the goal slower because vehicle 1 probably changes right then exits right and vehicle 2 exits right.''} \\
         & \textit{``likely reached the goal slower and with more jerk because vehicle 1 probably changes right then exits right.''} \\
         & \textit{``likely reached the goal slower and with more jerk and with less angular acceleration because vehicle 1 probably changes right then exits right.''}
    \end{tabular}
    \caption{\textit{(Scenario S1)} Varying number of causes and effects for the counterfactual $\vect{\omega}^{1}_{CF}=[\text{}\textit{Continue}]$. Including more information in the explanation improves its informativity, however longer explanations become more difficult to comprehend.}
    \label{tab:s1_expl_more}
\end{table*}

The criteria for human-centric AI set out in Section~\ref{sec:intro} necessitate our system to be transparent, causal, and intelligible.
Our system is transparent by design, as neither IGP2 nor XAVI rely on any components that are black boxes or otherwise uninterpretable.
We would then like to understand how well XAVI can capture the causal relationships when tested in realistic driving scenarios, and would also like to assess the intelligibility of our generated explanations. 

For this, we perform a preliminary evaluation of the XAVI system in two simulated driving scenarios powered by the high-fidelity, open-source CARLA~\cite{dosovitskiyCARLAOpenUrban2017} simulation environment.
The scenarios used here are scenarios S1 and S2 from the evaluation section of IGP2. 
\citeauthor{albrechtInterpretableGoalbasedPrediction2021} give intuitive and rational explanations about the behaviour of the ego vehicle for each scenario presented in the IGP2 paper. 
In particular, details of scenarios S1 and S2 and the explanations of the observed behaviours are presented in Figure~\ref{fig:scenarios}.
We rely on these explanations as ground truth to see how well our generated explanations \textit{match the causal attributions} of the ground truth explanations. 
We also vary the number of effects and causes passed to the explanation generation grammar to assess how the intelligibility of generated explanations changes with the complexity of the explanations.

To increase the diversity of our explanations, we perform ten simulations for each scenario where we randomly initialise the positions of the vehicles around their pre-defined starting points in a 10 meters longitudinal range. We also randomly initialise all vehicles' velocities in the range $[5,10]$ m/s. 
In all scenarios, the counterfactual action specifies the value for the first macro action selection random variable $\Omega_1$.
So for example, the query \textit{``Why did you change left instead of continuing straight?''} corresponds to $\Omega_1=\vect{\omega}_{CF}=[Continue]$. 

For scenario S1, we test two counterfactual actions: in one the ego continues straight behind vehicle 1 until it reaches its goal, so that $\vect{\omega}^{1}_{CF}=[\text{\textit{Continue}}]$; in the other, the ego turns right at the junction, so $\vect{\omega}^{1}_{CF}=[\text{\textit{Exit-right}}]$\footnote{
The macro action \textit{Exit-right} is a sequence of three manoeuvres: it encodes lane following until the junction, giving right-of-way, and turning. This means that ego will follow behind vehicle 1 also when executing \textit{Exit-right}.}.

For scenario S2, we test the following two counterfactual actions: $\vect{\omega}^{2}_{CF}=[\text{\textit{Exit-left}}]$, and $\vect{\omega}^{2}_{CF}=[\text{\textit{Exit-straight}}]$.
Note, that \textit{``Exit-straight''} is used here to differentiate the action from regular \textit{``Continue''} as the former macro action encodes giving way at a junction while the latter does not.

The generated explanations which include at most a single cause and a single effect are shown in Tables~\ref{tab:s1_expl} and \ref{tab:s2_expl}.
Our system is able to correctly identify the effects of switching to the counterfactuals while also explaining which actions of the other vehicles (if any) caused those effects.
In scenario S1, XAVI also revealed a possible collision outcome when the ego and vehicle 1 are spawned close to one another. 
Note, this outcome did not occur in the original IGP2 paper due to differences between random initialisations of vehicle positions.

Explanations with more than one cause or effect for scenario S1 are shown in Table~\ref{tab:s1_expl_more}.
We do not have a similar table for scenario S2 as all relevant causal information can be captured by at most one cause. 
This is because the actions of vehicle 2 do not affect the actions of the ego directly in any way.
For scenario S1, we can see that the shorter explanations can already capture the most crucial causal information of the ground truth, but more effects can be uncovered by XAVI.
However, more detailed explanations increase the complexity of explanations which may make them harder to understand.

\section{Discussion}\label{sec:discussion}

Our results show that XAVI successfully captures some of the causal relationships as compared with the ground truth explanations from IGP2, while also being able to discover other, unexpected outcomes.
The system is then able to generate intelligible explanations of varying complexity.

However, there are limitations to our work that need to be addressed in future work.
One limitation of the method is its inability to explain the causes behind actions of traffic participants in terms of properties that are lower level than macro actions, e.g. features of raw trajectories.
We would like to be able to justify our actions with causes that are finer in detail than the very high-level macro actions we currently have.
High-level macro actions may encode different behaviours depending on how the other traffic participants are acting, therefore formulating causes in terms of macro actions may mask crucial differences between different runs of simulation.
Indeed, the given causes for the ego's actions in Tables~\ref{tab:s1_expl} and \ref{tab:s1_expl_more} were the same between counterfactuals, but it is clear that for different counterfactuals different causes relating to the particular motion of vehicles would be more relevant.
For example, in Table~\ref{tab:s1_expl} for the counterfactual \textit{Continue} where ego reaches its goal slower, we should mention that vehicle 1 is \textit{slowing down} for a turn instead of just mentioning that it is exiting right.
On the other hand, for the same counterfactual where ego collides with vehicle 1 the more relevant cause for the collision is the actual fact that vehicle 1 unexpectedly changes right.
To enable this lower-level extraction of causes would mean that we need to find a way to compare and filter features of trajectories based on their causal relationships to other variables, which is a difficult task given the complexity of driving environments.

Automatic explanation generation methods are by their very nature \textit{post-hoc}, that is they work after our decisions were made. 
A common concern with any such post-hoc method is that they may not be sound, that is, faithful to the workings of the system they are explaining.
Without a formal proof of soundness, we cannot fully claim that XAVI is totally faithful to IGP2.
For example, XAVI does not represent each time step of the simulations explicitly or reason about how Q-values are updated.
However, given the variables XAVI \textit{does} reason about, we argue that our model is constructed to follow the steps of MCTS exactly without changing, removing, or adding extra information over a completed IGP2 planning run.

Another aspect to consider relates to the queries of passengers.
While contrastive explanations work well for \textit{why}-questions, there are many other types of questions users may ask (e.g. \textit{``What?''}, \textit{``How?''}), and we should support these lines of queries in the future.
Parsing passengers' questions is also a non-trivial task.
How could we know which macro actions a passenger is referring to in their question? 
What if those macro actions are not at all in our search tree? 
This last question also shows that we need a principled way to deal with missing data or cases where the system cannot give an explanation.
What is more, giving explanations where algorithmic exceptions are present in a non-misleading and consistent way is especially important, as these explanations reveal shortcomings of our systems, that may reduce trust levels.

One aspect of human-centric AI, which we did not mention in this work is the benefit of being dialogue-oriented. 
\citeauthor{millerExplanationArtificialIntelligence2019}~(\citeyear{millerExplanationArtificialIntelligence2019}) strongly argues that human-centric systems should be able to hold conversations with their human partners and allow opportunities for users to pose follow-up questions.
This is beneficial for the users because they can converse with our system as long as their curiosity or information-need is not satisfied.
Moreover, the system itself benefits from being able to hold conversations, as it can put the system on equal social status with humans, which is fundamental for developing trust~\cite{largeSteeringConversationLinguistic2017}.
We may also use follow-up questions to assess the passengers' understanding of our system, and deliver relevant explanations that are specifically designed to match the individual needs of each passenger.

Our evaluation of XAVI is preliminary, though the results are encouraging.
However, we need to test our system on many more interesting scenarios so that we can generate a more varied set of explanations if we want to be certain that XAVI does indeed work properly and is useful for passengers.
Besides the scenarios by~\citeauthor{albrechtInterpretableGoalbasedPrediction2021}~(\citeyear{albrechtInterpretableGoalbasedPrediction2021}), we can base further evaluation on the scenarios presented by~\citeauthor{wiegandExplanationThatExploring2020}~(\citeyear{wiegandExplanationThatExploring2020}) which were specifically collected to evaluate XAI in autonomous driving scenarios.
In the future, it will also be important to run a user study on how the generated explanations affect trust and knowledge levels in humans, as our ultimate goal is to achieve trustworthy autonomous driving.
Moreover, this will help to quantitatively assess the intelligibility of generated explanations and compare XAVI to other explanation generation systems.

\section{Conclusion}\label{sec:conclusion}
In this paper, we present an explanation generation system called eXplainable Autonomous Vehicle Intelligence (XAVI).
XAVI is designed to be fully transparent, causal, and intelligible thereby building towards a more human-centric explainability approach.
It is based on mapping a Monte Carlo Tree Search-based motion planning and prediction system for autonomous vehicles to a Bayesian network that models causal relationships in the planning process.
Preliminary evaluation of the system on a driving scenario shows that XAVI can accurately retrieve the causes behind and the effects of an autonomous vehicle's actions, and generate intelligible explanations based on causal information.
We also discuss several possible next steps and issues that need to be addressed in future work, such as lower-level causes, conversation-enabled systems, the need for error-handling, and question parsing.

\section*{Acknowledgements}
The authors would like to thank Cillian Brewitt and the anonymous reviewers for their helpful feedback.
This work was supported in part by the UKRI Centre for Doctoral Training in Natural Language Processing, funded by the UKRI (grant EP/S022481/1) and the University of Edinburgh, School of Informatics and School of Philosophy, Psychology \& Language Sciences.

\bibliographystyle{named}
\bibliography{ijcai22}

\clearpage
\appendix

\clearpage

\begin{figure*}
    \centering
    \begin{alignat*}{2}
        & S[s,\vect{e},\vect{c}] &&\rightarrow\text{ if }ACTION[\varepsilon, s.\vect{\omega}, \varnothing]\text{ then }EFFECTS[s.o, s.p,\vect{e}]\text{ because }CAUSES[\vect{c}]\text{ . } \\
        & ACTION[i, \vect{\omega}, p] &&\rightarrow str(i) \: ADV[p] \text{ }MACROS[\vect{\omega}] \\
        & MACROS[\vect{\omega}] &&\rightarrow str(\vect{\omega})_{|\vect{\omega}|=1} \: | \: MACROS[\vect{\omega}_1]\text{ then }MACROS[\vect{\omega}_{2:}] \\
        & EFFECTS[o,p,\vect{e}] &&\rightarrow \text{it would have } OUT[o,p]\:\: COMPS[\vect{e}] \\
        & COMPS[\vect{e}] &&\rightarrow \epsilon_{\vect{e}=\varnothing}\:\: |\:\: COMP_{|\vect{e}|=1}[\vect{e}]\:\: |\:\: COMPS[\vect{e}_1] \text{ and } COMPS[\vect{e}_{2:}] \\
        & COMP[e] &&\rightarrow \text{with }REL[e.\delta] \:\: str(e.r) \\
        & CAUSES[\vect{c}] &&\rightarrow \epsilon_{\vect{e}=\varnothing}\:\: |\:\:CAUSE_{|\vect{c}|=1}[\vect{c}]\: |\: CAUSES[\vect{c}_1]\text{ and }CAUSES[\vect{c}_{2:}] \\
        & CAUSE[\vect{c}] &&\rightarrow ACTION[c.i,c.\vect{\omega},c.p]  \\
        & OUT[o,p] &&\rightarrow ADV[p]\: str(o) \\
        & REL[\delta] &&\rightarrow \text{lower}_{\delta<0}\: | \:\text{higher}_{\delta>0}\: |\: \text{equal}_{\delta=0} \\
        & ADV[p] &&\rightarrow \text{never}_{p=0}\: |\: \text{unlikely}_{0<p\leq0.33}\: |\: \text{probably}_{0.33<p\leq0.67} | \\
        & &&\quad\:\:\,\text{likely}_{0.67<p<1.0}\: |\: \text{certainly}_{p=1.0}\: |\: \epsilon_{p=\varnothing}
    \end{alignat*}
    \caption{The explanation generation grammar rules. The function $str(.)$ returns a pre-defined textual representation of its argument. Subscripts denote conditions for the rule to be applicable. Note, $\epsilon$ denotes the empty string while $\varepsilon$ the ego vehicle.}
    \label{fig:cfg}
\end{figure*}

\end{document}